# Design and Implementation of Image Processing System for Lumen Social Robot-Humanoid as an Exhibition Guide for Electrical Engineering Days 2015

Setyaki Sholata Sya[1], Ary Setijadi Prihatmanto[2]

#School of Electrical Engineering and informatics, Institut Teknologi Bandung
Jalan Ganesha 10, Bandung 40132, Indonesia
[1]`setyaki.s.s@gmail.com`
[3]`asetijadi@lskk.ee.itb.ac.id`

*Abstract*— Lumen Social Robot is a humanoid robot development with the purpose that it could be a good friend to all people. In this year, the Lumen Social Robot is being developed into a guide in the exhibition and in the seminar of the Final Exam of undergraduate and graduate students in Electrical Engineering ITB, named Electrical Engineering Days 2015. In order to be the guide in that occasion, Lumen is supported by several things. They are Nao robot components, servers, and multiple processor systems. The image processing system is a processing application system that allows Lumen to recognize and determine an object from the image taken from the camera eye. The image processing system is provided with four modules. They are face detection module to detect a person's face, face recognition module to recognize a person's face, face tracking module to follow a person's face, and human detection module to detect humans based on the upper parts of person's body. Face detection module and human detection module are implemented by using the library harcascade.xml on EMGU CV. Face recognition module is implemented by adding the database for the face that has been detected and store it in that database. Face tracking module is implemented by using the Smooth Gaussian filter to the image.

## I. INTRODUCTION

Robot has grown significantly both in function and form. Not only in the world of industry, robots are also developed to be the human's friend as a social robot. According to Hegel et al, a robot can be called a social robot if it has an appearance and social function [1]. To be able to have an appearance and social function, social robot needs to have a shape resembling a human body structure commonly which is commonly called a humanoid robot. So the Lumen robot which is one example of a social-humanoid robot can be used as a guide in both indoors and outdoors condition.

This paper will explain the implementation and design of an image processing system in the development of a Lumen robot as a guide at an exhibition and seminar of Final undergraduate and graduate students of Electrical Engineering ITB, namely Electrical Engineering Days 2015. Through this image processing system, Lumen will be able to detect people, know the people and recognize them.

## II. ANALYSIS AND DESIGN SYSTEM

### A. Robot Nao

Lumen robot is a humanoid Nao robot manufactured by the French company named Aldebaran Robotics. Nao has the look of a child with the height of 573 mm and weight of 4.996 kg. Nao has two autofocus camera located on the forehead and mouth and the camera has the ability of 30 fps, 640 * 480 pixels and has a maximum focus of 6 m. Nao also has an API that can process image processing, especially for the detection of a person's face. The API module named ALFaceDetection. The module can also store data of the detected faces to Nao's memory in order to Nao can recognize the faces and perform the repeating detection.

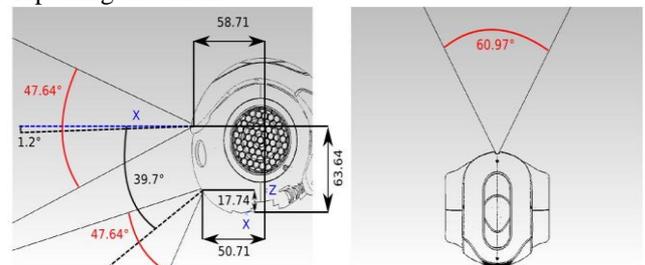

Figure 1. Spesification of Nao's Camera

Image processing system in Lumen does not use API Nao because of the limited ability in image processing in API and the limited internal memory in saving the face database through this API command.

### B. Emgu CV

Emgu CV is a cross-platform image processing library. Emgu CV is closely related to OpenCV because Emgu CV is a .NET wrapper for OpenCV or it can be said that Emgu CV is OpenCV in .NET. The language of the program in Emgu CV is C #, VB, IronPython and VC ++. Emgu CV can also be used in Linux, Windows, Mac OS X, and various types of mobile as Android, iPhone, iPod Touch, and iPad.

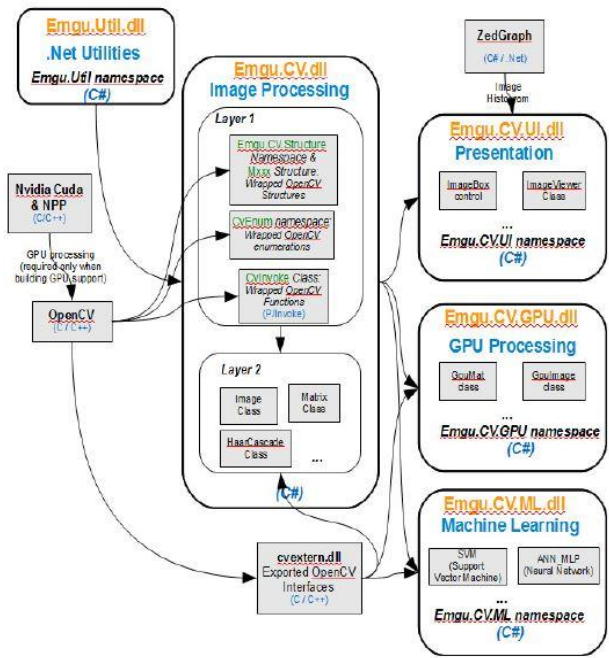

Figure 2 Platform Emgu CV

One advantage of using Emgu CV in performing image processing is that there are lots of library xml that are much related to image processing. The Xml library used in this detection is haar_cascade_face.xml for detecting people's faces and haar_cascade_upperbody.xml to detect the people.

*C. Haar Cascade*

Face detection is based on the identifying and finding the location of the human face image in a picture regardless of the size, position, and condition (Padmaja & Prabakar, 2012). This also applies in the detection of a person based on the detection of the upper body.

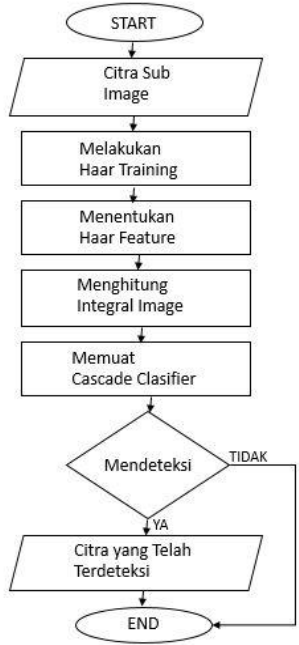

Figure 3 Flow Chart Method Haar Cascade

Haar cascade has 4 main concepts in conducting the detection, namely Haar Training, Haar Feature, Integral Image, and Cascade Clasifier.

Based on Figure 3, the image that we want to detect will be tested using the haar training. The purpose from haar training is to separate the object that we wish to detect with the object that we do not want to detect by making the positive and negative samples. After that, the next process will be the haar feature process, which will decide whether there is an object in the image or not by calculating the difference from the total of black pixels with the total of white pixels. There are four types of features based on the number of rectangles (Krishna & Srinivasulu, 2012).

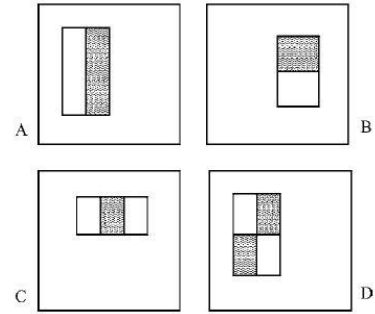

Figure 4 Haar Feature

Haar feature will search for the position of the object by looking for features that have a high degree of differentiation. This is obtained from the threshold value which is the result of the difference of the black and white pixels.

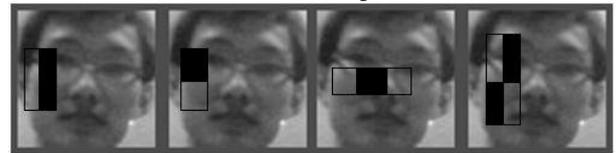

Figure 5 Search by Haar Feature

Once the face is detected, Integral Image will be conducted. Integral image is the detection with a more efficient and more number of haar feature.

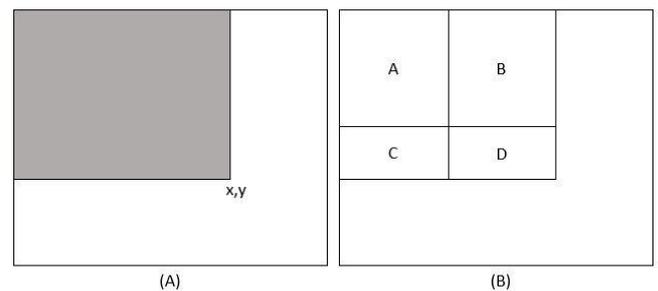

Figure 6 Integral Image Method

$$ii(x, y) = \sum_{x' \leq x, y' \leq y} i(x', y')$$

Haar cascade is a learner and has a weak classifier, so the work of haar cascade should be done massively. The more the haar cascade process, the more accurate the result is. This many Haar cascade features is organized by the cascade classifier.

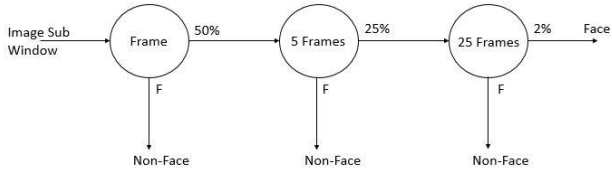

Figure 7 Cascade Classifier

When there is only one frame being detected, the error was obtained by 50%. But when the detection is using many frames (minimumly 24 frames), the error was only 2%.

### D. EigenFace

Eigenface is a set of face standardize ingredient derived from statistical analysis of many faces (Layman in Al Fatta, Hanif, 2009). Eigenface is one method in recognizing faces using the calculation from Principal Component Analysis (PCA). PCA itself is a technique to reduce the dimensions of a space presented by statistical variable xn where the variable is correlated with each other. The intention of using PCA in face recognition is to form a face room by finding the eigenvector and eigenvalue that corresponds to the largest eigen values from the face image.

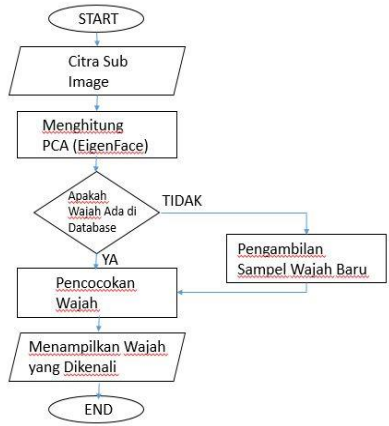

FIgure 8 Flow Chart EigenFace Method

Each face image that is stored on the database has values that will be used in the calculation of eigenvalue and eigenvector PCA in order to recognize that face image. PCA calculation is derived from the eigenvalue and eigenvector. Eigenvalue is the transformation of each pixel in the image into a column vector that will produce a matrix. While eigenvector is the covariance of that matrix. Then, each value that has been recorded will be summed and divided by the number of face experiments conducted. This value is used as a reference values.

$$\mu = \frac{1}{n}\sum_{i=1}^{n} xi$$

$$S = \frac{1}{n}\sum_{i=1}^{n}(xi-\mu)(xi-\mu)^T$$

$$Sv_i = \lambda_i v_i \ ; \ i = 1, 2, 3, ..., n$$

μ = mean
S = covarian
$v_i$ = eigenvector
λ = eigenvalue

When it detects a new face, eigenvalue will get a new value. Then the new value is compared with eigenvalue in the database. If the value of the new eigenvalue is higher than the old eigenvalue, then the face of the person is detected. Whereas if not, then that person is not detected.

### E. Gaussian Smoothing Filtering

The method used in following the detected face is Smooting Gaussian filter. This method performs a filter on an image and focus it on the desired object.

Initially, the image will be normalized in the RGB color. The purpose of the normalization is to make the range of the image value between 0-225. Once normalized, RGB is converted to HSV. The aim from the conversion is to get the threshold value in detecting a face that will be tracked. By giving maximum and minimum limit value, value of hue, saturation and value at one point pixel of the object's surface can be obtained. Pixels colour that have the value in the threshold will be changed to white, while outside of the threshold will be changed to black. However, at this stage, the results still have a lot of noise there.

To get a smoother detection and to eliminate the noise, the Gaussian Smoothing Filter process needs to be done.

## III. IMPLEMENTATION

### A. Face Detection

The experiment of the face detection application was conducted 60 times in the morning, afternoon, evening, and night with three different lighting conditions (dark, adequate light, and excessive light). Each lighting was also done in five different positions of the object.

In dark condition, the detection is not working. As for the adequate and excessive light, the detection can work but the performance of the detection decreases when in excessive light. This is shown by the shrinking of the displacement position angle when it gets more light.

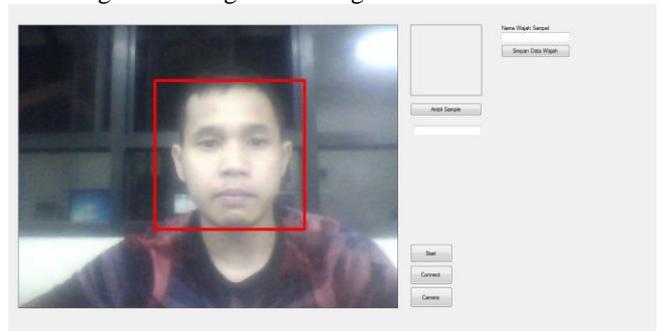

Figure 9 The results of the face detection

## B. Face Recognizion

The experiments of the face recognition application is done by taking three different faces with each is stored as many as 60 pieces of the faces. At trial, face recognition error was occurred in detecting the people's face. This is because the value of the eigenvalue from the face that we want to recognize is greater than the value of the eigenvalue of the face that is stored in the database. When it happens, the application will recognize that person's face as the face that has been stored in the database. Besides that, the lighting, the distance and angle of the camera view can affect the person's face recognition.

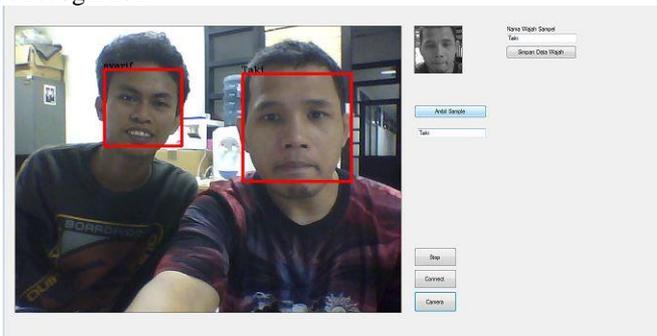

Figure 10 Results of face recognition

## C. Face Tracking

Face tracking is the first step in doing face tracking in the module motion. Face tracking has managed to lock the face that will be tracked with a green dot in the center of the red square.

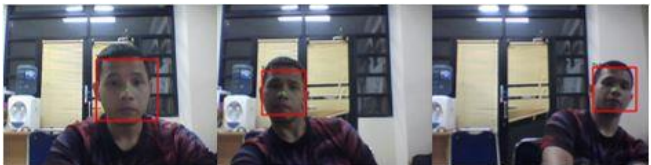

Figure 11 Results of the face tracking

## D. Human Detection

The application of the human detection is almost the same as the face detection. The experiments were carried out 60 times in the morning, afternoon, evening, and night with three different lighting conditions (dark, adequate light, and excessive light). Each lighting was also done in five different positions of the object. The performance of human detection also decreases as the face detection experiments.

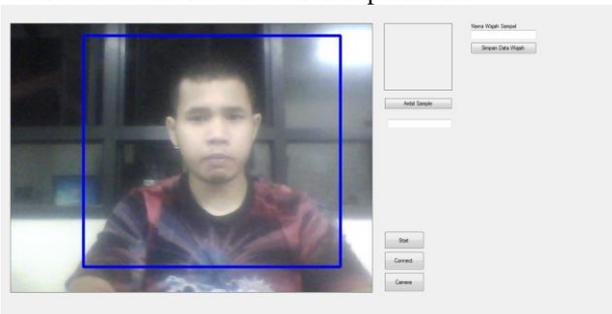

Figure 12 The result of the detection of people based on his upper body

## IV. CONCLUSION

The conclusion from the implementation that has been done is

1. By using the applications in image processing system modules (human detection, face detection, face recognition, and face tracking), Lumen has been able to act as a guide of the exhibition in Electrical Engineering Days 2015 quite satisfactory.
2. In image processing, lighting, perspective, and visibility is the factors that influence the treatment process.
3. In recognizing a person's face, the more samples that is being stored, the more accurate the detection of person's identity will be but it can reduce the performance in recognizing him.

## ACKNOWLEDGMENT

Lumen development project is funded by the Ministry of Education and Culture of Indonesia with the support from the Laboratory of Computer Control and System Institut Teknologi Bandung (ITB)

# Desain dan Implementasi Sistem Pengolahan Citra untuk Lumen Robot Sosial Humanoid sebagai Pemandu Pameran pada *Electrical Engineering Days* 2015

Setyaki Sholata Sya[1], Ary Setijadi Prihatmanto[2]

#*Sekolah Teknik Elektro dan Informatika, Institut Teknologi Bandung*
*Jalan Ganesha 10, Bandung 40132, Indonesia*
[1]`setyaki.s.s@gmail.com`
[3]`asetijadi@lskk.ee.itb.ac.id`

*Abstract*— **Lumen Sosial Robot merupakan sebuah pengembangan robot humanoid agar dapat menjadi teman bagi banyak orang. Pada tahun ini, Lumen Sosial Robot dikembangkan menjadi pemandu pameran pada suatu pameran dan seminar Tugas Akhir mahasiswa sarjana dan pascasarjana Tenik Elektro ITB, yaitu *Electrical Engineering Days* 2015. Agar dapat menjadi pemandu pameran, Lumen didukung beberapa hal, yaitu komponen robot Nao, server, dan beberapa sistem pengolah. Sistem pengolahan citra merupakan sistem aplikasi pengolah yang bertujuan Lumen dapat mengenali dan mengetahui suatu objek pada citra yang diambil dari camera mata Lumen. System pengolahan citra dilengkapi dengan empat buah modul, yaitu modul *face detection* untuk mendeteksi wajah seseorang, modul *face recognition* untuk mengenali wajah orang tersebut, modul *face tracking* untuk mengikuti wajah seseorang, dan modul *human detection* untuk mendeteksi manusia berdasarkan bagian tubuh atas orang tersebut. Modul *face detection* dan modul *human detection* diimplementasikan dengan menggunakan library harcascade.xml pada EMGU CV. Modul *face recognition* diimplementasikan dengan menambahkan database untuk wajah yang telah terdeteksi dan menyimpannya pada database. Module *face tracking* diimplementasikan dengan menggunakan filter Smooth Gaussian.**

*Keywords*— **Lumen, system pengolahan citra,** *face detection, face recognition, face tracking, human detection.*

## I. Pendahuluan

Robot telah berkembang secara pesat baik secara fungsi maupun bentuk. Tidak hanya di dunia industri, robot juga dikembangkan untuk menjadi teman dan sahabat manusia, seperti robot sosial. Menurut Hegel et al, suatu robot dapat disebut sebagai robot sosial jika memiliki fungsi dan tampilan sosial [1]. Agar dapat memiliki fungsi dan tampilan sosial itu, robot sosial memiliki bentuk menyerupai struktur tubuh manusia yang biasa disebut robot humanoid. Sehingga robot Lumen yang merupakan salah satu contoh robot sosial-humanoid dapat dimanfaatkan sebagai robot pemandu baik dalam keadaan indoor ataupun outdoor.

Pada makalah ini akan dijelaskan tentang implementasi dan desain suatu sistem pengolahan citra pada pengembangan robot Lumen menjadi pemandu pameran pada suatu pameran dan seminar Tugas Akhir mahasiswa sarjana dan pascasarjana Teknik Elektro ITB, yaitu *Electrical Engineering Days* 2015. Melalui sistem pengolahan citra ini, Lumen mendeteksi orang, mengetahui orang dan mengenali orang tersebut.

## II. Analisis dan Sistem Desain

### A. Robot Nao

Robot Lumen merupakan robot Nao berjenis *humanoid robot* yang diproduksi oleh perusahan di Prancis yang bernama Aldebaran Robotics. Nao memiliki tampilan seperti anak kecil yang memiliki tinggi 573 mm dan berat 4,996 kg. Nao memiliki dua buah kamera autofocus yang terletak pada dahi dan mulutnya dan kamera memiliki kemampuan 30 fps, 640*480 pixel dan memiliki titik focus maksimal 6 m. Nao juga memiliki API yang dapat memproses pengolah citra terutama untuk pendeteksian wajah seseorang. Modul API tersebut bernama ALFaceDetection. Pada modul tersebut juga dapat menyimpan data wajah yang telah terdeteksi ke memory Nao guna Nao dapat mengenali wajah tersebut dan melakukan pengulangan pendeteksian itu.

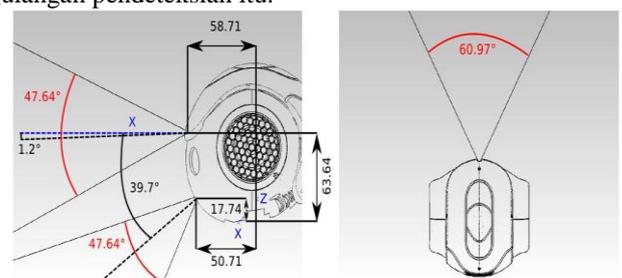

Gambar 1 Spesifikasi Kamera Nao

Sistem pengolahan citra pada Lumen tidak menggunakan API Nao karena adanya keterbatasan kemampuan dalam pengolahan citra API dan keterbatasan memory internal saat menyimpan database wajah melalui perintah API ini.

### B. Emgu CV

Emgu CV merupakan suatu lintas platform library pengolahan citra. Emgu CV berkaitan erat dengan OpenCV

karena Emgu CV merupakan pembungkus NET untuk OpenCV atau bisa dibilang Emgu CV adalah OpenCV di NET. Bahasa program yang terdapat pada Emgu CV ialah C#, VB, IronPython dan VC++. Emgu CV juga dapat digunakan di Linux, Windows, Mac OS X, dan berbagai jenis mobile seperti Android, iPhone, iPod Touch, dan iPad.

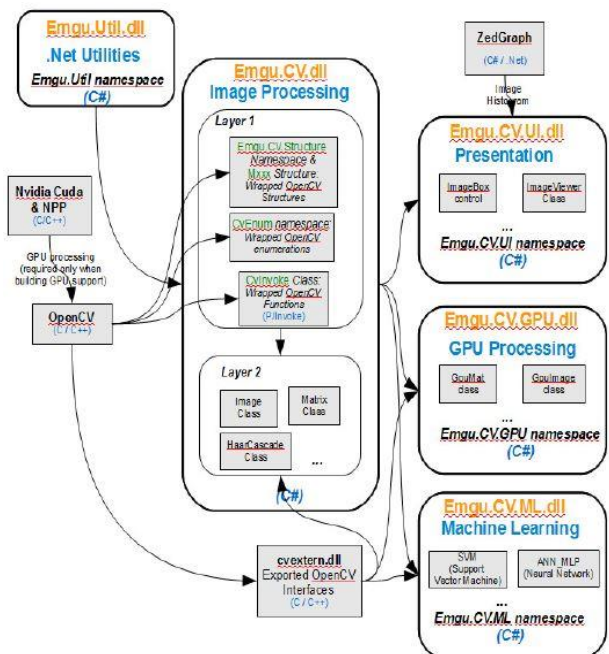

Gambar 2 Platform Emgu CV

Salah satu keuntungan menggunakan Emgu CV untuk melakukan pengolahan citra adalah banyak library xml yang berkaitan dengan pengolahan citra. Library xml yang dipakai dalam pendeteksian ini adalah haar_cascade_face.xml untuk pendeteksian wajah orang dan haar_cascade_upperbody.xml untuk mendeteksi orang.

## C. Haar Cascade

Deteksi wajah didasarkan pada identifikasi dan menemukan lokasi citra wajah manusia dalam gambar terlepas dari ukuran, posisi, dan kondisi (Padmaja & Prabakar, 2012). Hal ini juga berlaku dalam pendeteksian seseorang berdasarkan pendeteksian tubuh bagian atas.

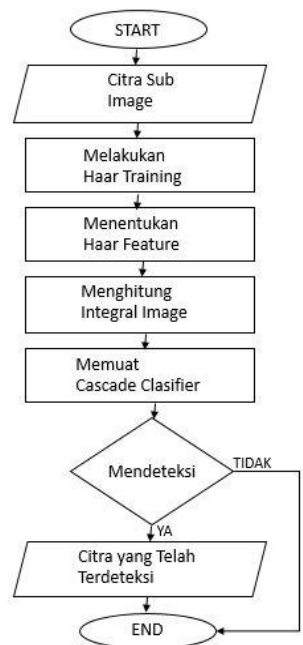

Gambar 3 Flow Chart Metode Haar Cascade

Haar Cascade memiliki 4 konsep utama dalam melakukan pendeteksian, yaitu Haar Training, Haar Feature, Integral Image dan Cascade Clasifier.

Berdasarkan Gambar 3, citra yang akan dideteksi akan di-uji menggunakan *haar training*. Tujuan melakukan *haar training* adalah memisahkan objek yang ingin dideteksi dengan objek yang tidak ingin dideteksi dengan cara membuat *sample positive* dan *negative*. Setelah itu akan dilakukan proses *haar feature*, yaitu memutuskan apakah di citra tersebut terdapat objek atau tidak dengan cara melakukan pengurangan terhadap akumulasi piksel hitam dengan akumulasi piksel putih. Terdapat empat jenis fitur berdasarkan jumlah persegi panjang (Krishna & Srinivasulu, 2012).

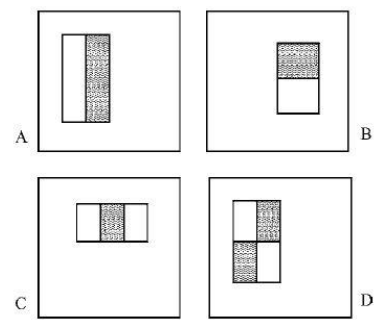

Gambar 4 Haar Feature

*Haar feature* akan melakukan pencarian posisi objek dengan cara mencari fitur-fitur yang memiliki tingkat pembeda yang tinggi. Tingkat pembeda ini diperoleh dari nilai ambang (threshold) yang merupakan hasil selisih dari piksel hitam dan putih tersebut.

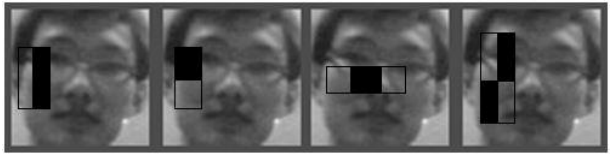

Gambar 5 Pencarian oleh Haar Feature

Setelah terdeteksi wajah, maka akan dilakukan *Integral Image*. *Integral Image* adalah pendeteksian dengan jumlah haar feature yang lebih banyak dan efisien.

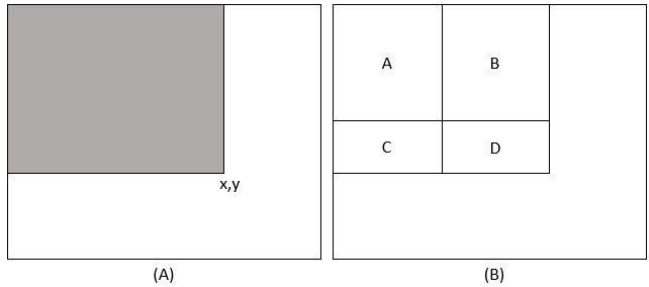

Gambar 6 Metode *Integral Image*

$$ii(x,y) = \sum_{x' \leq x, y' \leq y} i(x', y')$$

*Haar cascade* memiliki sifat *learner* dan *classifier* yang lemah, sehingga pengerjaan *haar cascade* harus dilakukan secara masal. Semakin banyak proses *haar cascade*, maka hasil yang diinginkan akan semakin akurat. Proses *haar feature* yang banyak ini diorganisir oleh *cascade classifier*.

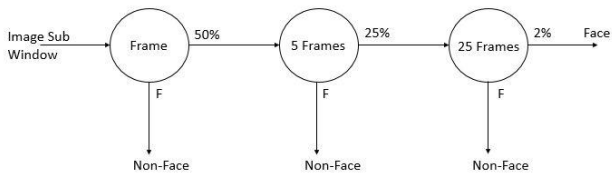

Gambar 7 Cascade Classifier

Ketika hanya satu frame yang dideteksi, error yang diperoleh sebesar 50%. Namun dengan melakukan pendeteksian dengan banyak frame (minimum 25 frames), error yang dihasilkan hanya 2%.

D. EigenFace

*EigenFace* adalah sekumpulan *standardize face ingredient* yang diambil dari analisis statistik dari banyak wajah (Layman dalam Al Fatta, Hanif, 2009). *EigenFace* merupakan salah satu metode dalam melakukan pengenalan terhadap suatu wajah dengan penghitungan *Principal Component Analysis* (PCA). PCA sendiri merupakan teknik untuk mengurangi dimensi sebuah ruang yang dipresentasikan oleh variable statistic $x_n$ di mana variable tersebut saling korelasi dengan lainnya. Tujuan penggunaan PCA pada pengenalan wajah adalah membentuk ruang wajah dengan cara mencari eigenvector dan eigenvalue yang berkoresponden dengan nilai eigen terbesar dari citra wajah.

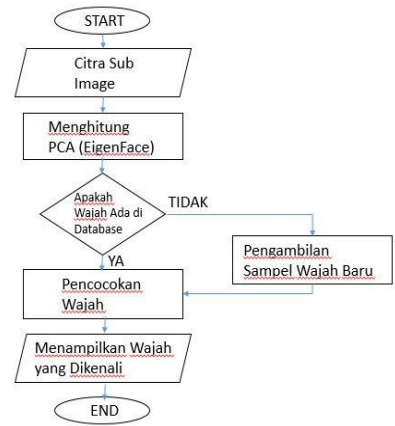

Gambar 8 Flow Chart metode EigenFace

Setiap citra wajah yang disimpan pada database memiliki nilai yang dipakai dalam penghitungan eiugenvalue dan eigenvector PCA guna mengenali citra wajah tersebut. Perhitungan PCA berasal dari eigenvalue dan eigenvector. Eigenvalue adalah tranformasi setiap piksel pada gambar menjadi vector kolom sehingga menghasilkan satu matriks. Sedangkan eigenvektor merupakan merupakan covarian matriks tersebut. Kemudian, setiap nilai yang dicatat, dijumlahkan dan dibagi dengan banyaknya percobaan wajah yang dilakukan. Nilai ini dijadikan sebagai nilai acuan.

$$\mu = \frac{1}{n}\sum_{i=1}^{n} xi$$

$$S = \frac{1}{n}\sum_{i=1}^{n}(xi-\mu)(xi-\mu)^T$$

$$Sv_i = \lambda_i v_i \,;\, i = 1, 2, 3, \ldots, n$$

$\mu$ = mean
$S$ = Kovarian
$v_i$ = eigenvector
$\lambda$ = eigenvalue

Ketika mendeteksi wajah baru, eigenvalue akan mendapat nilai baru. Kemudian nilai baru tersebut dibandingkan dengan eigenvalue pada database. Apabila nilai eigenvalue yang baru lebih besar dari eigenvalue yang lama, maka muka orang tersebut terdeteksi. Sedangkan jika tidak, maka orang tersebut tidak terdeteksi.

E. Gaussian Smoothing Filtering

Metode yang dipakai dalam mengikuti wajah yang terdeteksi ialah *Gaussian Smoting Filter*. Metode ini melakukan filter pada suatu citra dan memfokuskannya pasa objek yang diinginkan.

Awalnya, citra akan dinormalisasi pada warna RGB. Tujuan dari normalisasi agar *range* nilai citra berada diantara 0-225. Setelah dinormalisasi, RGB dikonversi menjadi HSV. Konversi ini dilakukan dengan tujuan untuk mendapatkan nilai threshold dalam mendeteksi wajah yang mau di-*tracking*. Dengan memberikan nilai batas maksimum dan minimum, maka akan mendapatkan nilai *hue*, *saturation* dan *value* pada satu titik piksel permukaan objek. Piksel warna yang memiliki nilai di dalam batas ambang akan diubah menjadi putih, sedangkan di luar akan menjadi hitam. Namun pada tahap ini, hasil yang didapat masih banyak terdapat noise.

Untuk mendapatkan pendeteksian yang lebih halus dan menghilangkan noise tersebut, maka dilakukan proses *Gaussian Smoothing Filter*.

### III. IMPLEMENTASI

#### A. Face Detection

Aplikasi *face detection* dilakukan percobaan sebanyak 60 kali pada pagi, siang, sore dan malam dengan 3 kondisi pencahayaan yang berbeda (gelap, cukup cahaya, dan cahaya berlebihan). Setiap pencahayaan juga dilakukan pada 5 posisi objek yang berbeda.

Pada kondisi gelap, pendeteksian tidak bekerja. Sedangkan untuk cukup dan berlebihan cahaya, pendeteksian bisa berjalan namun performansi deteksi menurun saat cahaya berlebih. Hal ini ditunjukkan dengan perubahan perpindahan posisi sudut yang mengecil ketika mendapat cahaya yang lebih banyak.

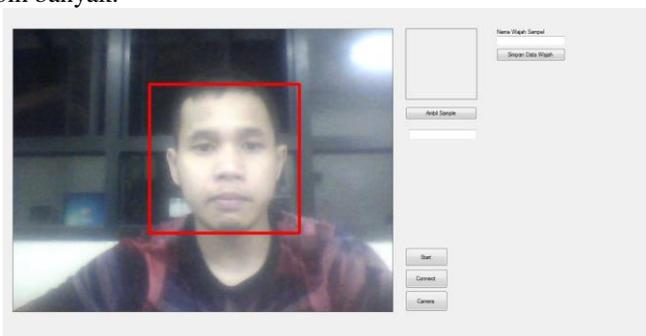

Gambar 9 Hasil pendeteksian wajah

#### B. Face Recognizion

Percobaan aplikasi face recognition dilakukan dengan pengambilan 3 wajah orang berbeda yang masing- masing disimpan sebanyak 60 buah sampel wajahnya. Pada percobaan, *face recognition* terjadi error seperti salah mendeteksi wajah orang. Hal ini disebabkan karena nilai eigenvalue wajah yang mau dikenali lebih besar dari pada nilai eigenvalue wajah yang terdapat di database. Bila terjadi hal itu, aplikasi akan mengenali wajah tersebut sebagai orang yang wajahnya telah terdapat pada database. Selain itu juga, pencahayaan, jarak dan sudut tampilan terhadap kamera bisa mempengaruhi pengenalan wajah orang tersebut.

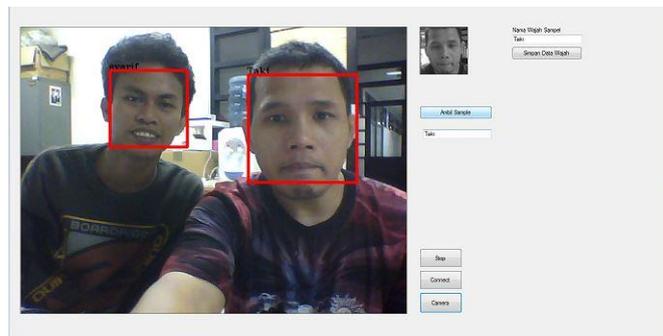

Gambar 10 Hasil pengenalan wajah

#### C. Face Tracking

*Face tracking* merupakan langkah awal dalam melakukan *face tracking* pada modul *motion*. *Face tracking* telah berhasil mengunci muka yang akan diikutinya dengan tanda titik warna hijau pada pusat persegi berwarna merah.

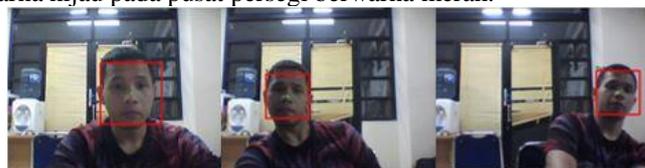

Gambar 11 Hasil tracking wajah

#### D. Human Detection

Aplikasi *human detection* hampir sama dengan *face detection*. Percobaan dilakukan sebanyak 60 kali pada pagi, siang, sore dan malam dengan 3 kondisi pencahayaan yang berbeda (gelap, cukup cahaya, dan cahaya berlebihan). Setiap pencahayaan juga dilakukan pada 5 posisi objek yang berbeda. Performasi pendeteksian manusia juga menurun seperti percobaan *face detection.*

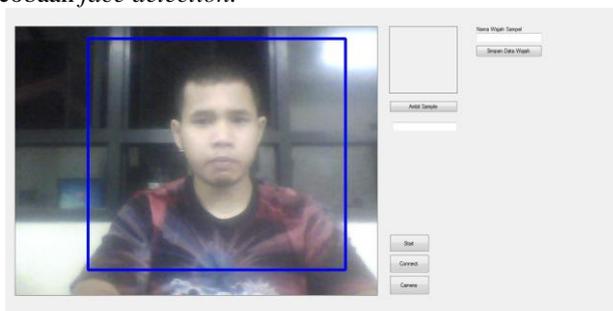

Gambar 12 Hasil pendeteksian orang berdasarkan tubuh bagian atasnya

### IV. KESIMPULAN

Kesimpulan dari implemtasi yang telah dilakukan adalah
1. Dengan menggunakan aplikasi-aplikasi pada modul sistem pengolahan citra (*human detection, face detection, face recognition,* dan *face tracking*), Lumen sudah dapat berperan menjadi pemandu pameran dalam *Electrical Engineering Days* 2015 dengan cukup baik.

2. Dalam pengolahan citra, pencahayaan, sudut pandang, dan jarak pandang memiliki faktor yang sangat mempengaruhi proses pengolahan tersebut.
3. Dalam pengenalan wajah seseorang, semakin banyak sampel yang disimpan maka semakin akurat pendeteksian akan identitas orang tersebut namun dapat menurunkan performasi dalam mengenalinya.

ACKNOWLEDGMENT